\pdfoutput=1

\documentclass[11pt]{article}

\usepackage{EMNLP2022}

\usepackage{times}
\usepackage{latexsym}

\usepackage[T1]{fontenc}

\usepackage[utf8]{inputenc}

\usepackage{microtype}

\usepackage{inconsolata}

\usepackage{amsmath}
\usepackage{amsfonts}
\usepackage{amssymb}
\usepackage{graphicx}
\usepackage{multirow}
\usepackage{booktabs}
\usepackage{diagbox}
\usepackage{bbding}
\usepackage{enumitem}
\usepackage{commath}
\usepackage{algorithm}
\usepackage{algorithmic}
\usepackage[toc,page]{appendix}
\newcommand{\bm}{\mathbf}

\usepackage{bm}
\usepackage{bbm}
\usepackage{multicol} 
\usepackage{amsthm}

\newcommand{\txw}{\textcolor{black}}
\newcommand{\gl}[1]{\textcolor{black}{#1}}
\newcommand{\gab}[1]{\textcolor{black}{#1}}
\newcommand{\jr}{\textcolor{black}}

\title{Event-Centric Question Answering via Contrastive Learning and Invertible Event Transformation}

\author{Junru Lu$^1$, Xingwei Tan$^1$, Gabriele Pergola$^1$, Lin Gui$^2$ and Yulan He$^{1,2,3}$ \\
  $^1$Department of Computer Science, University of Warwick, UK\\
  $^2$Department of Informatics, King's College London, UK\\
  $^3$The Alan Turing Institute, UK\\
    \texttt{\{Junru.Lu, Xingwei.Tan, Gabriele.Pergola\}@warwick.ac.uk} \\
  \texttt{\{lin.1.gui, yulan.he\}@kcl.ac.uk}}

\begin{document}
\maketitle
\begin{abstract}
  Human reading comprehension often requires reasoning of event semantic relations in narratives, represented by Event-centric Question-Answering (QA). 
  To address event-centric QA, we propose a novel QA model with contrastive learning and invertible event transformation, call \texttt{TranCLR}.  
  Our proposed model utilizes an invertible transformation matrix to project semantic vectors of events into a common event embedding space, trained with contrastive learning, and thus naturally inject event semantic knowledge into mainstream QA pipelines. The transformation matrix is fine-tuned with the annotated event relation types between events that occurred in questions and those in answers, using event-aware question vectors. Experimental results on the \textbf{E}vent \textbf{S}eman\textbf{t}ic R\textbf{e}lation \textbf{R}easoning (ESTER) dataset show significant improvements in both generative and extractive settings compared to the existing strong baselines,
  achieving over 8.4\% gain in the token-level F1 score and 3.0\% gain in Exact Match (EM) score under the multi-answer setting. 
  Qualitative analysis reveals the high quality of the generated answers by \texttt{TranCLR}, demonstrating the feasibility of injecting event knowledge into QA model learning. Our code and models can be found at  \url{https://github.com/LuJunru/TranCLR}.
\end{abstract}

\section{Introduction}

\begin{figure}[ht]
  \centering
  \includegraphics[width=\linewidth]{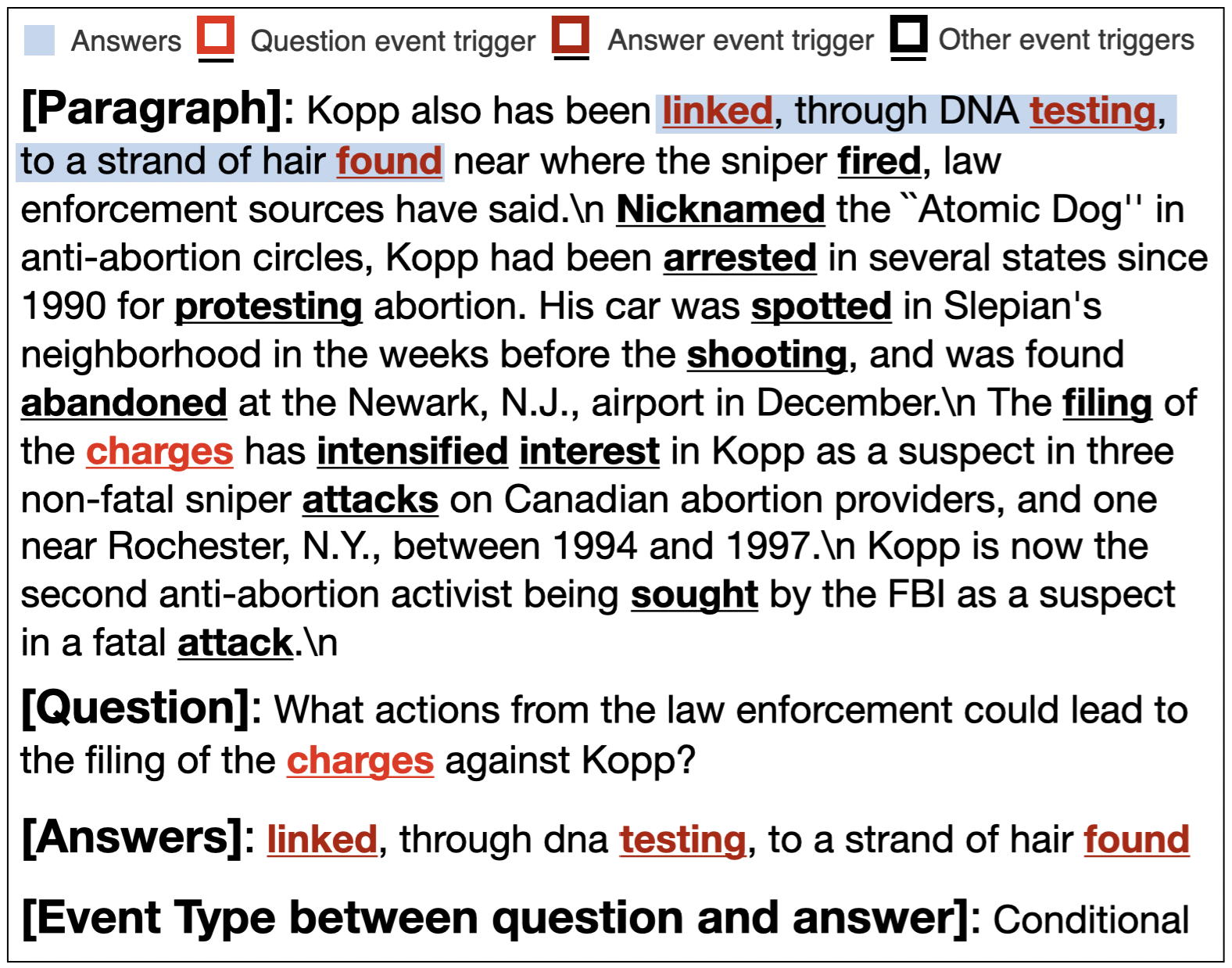}
  \caption{An event-centric QA example from the ESTER dataset \cite{han2021ester}. All event triggers are highlighted in bold and underlined in the paragraph. The question event trigger and answer event triggers are further highlighted in red colors with different shades. In-text answer is smeared with blue.}
  \label{fig:sample}
\end{figure}

Since 2019, many larger-scale pre-trained language models (PLMs) \cite{devlin2018bert,raffel2019exploring, lu2020chime, boost21} have been introduced to address the Question-Answering (QA) tasks, reaching performance on par with humans on entity-centric QA datasets such as SQuAD \cite{rajpurkar2016squad}, TriviaQA \cite{joshi2017triviaqa}, and NewsQA \cite{trischler2016newsqa}, in which answers are often entities extracted from text. A raising challenge is to research and develop new PLM-based frameworks tackling more difficult QA settings in real-world scenarios. 
One direction is to go beyond entity-centric QA and explore QA tasks focusing on high cognitive level information such as \textbf{events}. 
A recently introduced \textbf{E}vent \textbf{S}eman\textbf{t}ic R\textbf{e}lation \textbf{R}easoning (ESTER) dataset \cite{han2021ester} facilitates the development of Machine Reading Comprehension (MRC) models for event-centric QA. The dataset contains event-centric question-answers annotated with event semantic relation type labels. 
Figure \ref{fig:sample}\footnote{Better viewing in color.} shows an example instance from the dataset. The main challenge is to effectively explore event semantic knowledge to answer event-centric questions. In the example illustrated in Figure \ref{fig:sample}, an MRC or QA model needs to first understand that the question asks for a potential answer event which holds a \emph{conditional} relation with the main event `\emph{charges}' mentioned in the question. It then needs to identify events in the paragraph which have the \emph{conditional} relation with the question event trigger `\emph{charges}', in this case, `\emph{linked}', `\emph{testing}' and `\emph{found}'. Finally, it needs to generate the answer involving the identified events in natural language. It is easy for humans to understand narratives by constructing a situational logic chain capturing how events evolve and relate to each other in text. 
Yet, existing QA models only learn shallow semantic cues based on word token statistics gathered from large-scale text corpora \cite{niven2019probing}, but are not able to grasp high-level concepts such as events. Preliminary experimental results using the pre-trained T5 language model on the ESTER dataset show that there remains a large gap over 15\% between machine and human performance \cite{han2021ester}. 

\gl{Intuitively, it is possible to inject event information through a multi-task learning framework where event-related tasks such as event relation type detection and event embedding learning could be potentially useful to guide the QA model to generate better answers. For example, event relation type detection aims to detect the desired event semantic relation given a question, while event embedding learning aims to push events holding the desired semantic relation closer in the new event embedding space. However, in a QA model, the answer generator is usually built on a PLM in which the original event representations learned in the PLM should be preserved. That is, we want to map event representations onto a new event embedding space in order to inherently capture their semantic relations specified by an input question, but at the same time, we need to keep the original event representations learned by the PLM in order to generate coherent answers. To deal with this dilemma, 
we propose an invertible transformation operator, which makes it possible to learn new event embeddings
without changing the mutual information of any given event pairs, making it effective in injecting event information for event-centric QA. }

\begin{figure*}[ht]
  \centering
  \includegraphics[width=0.82\textwidth]{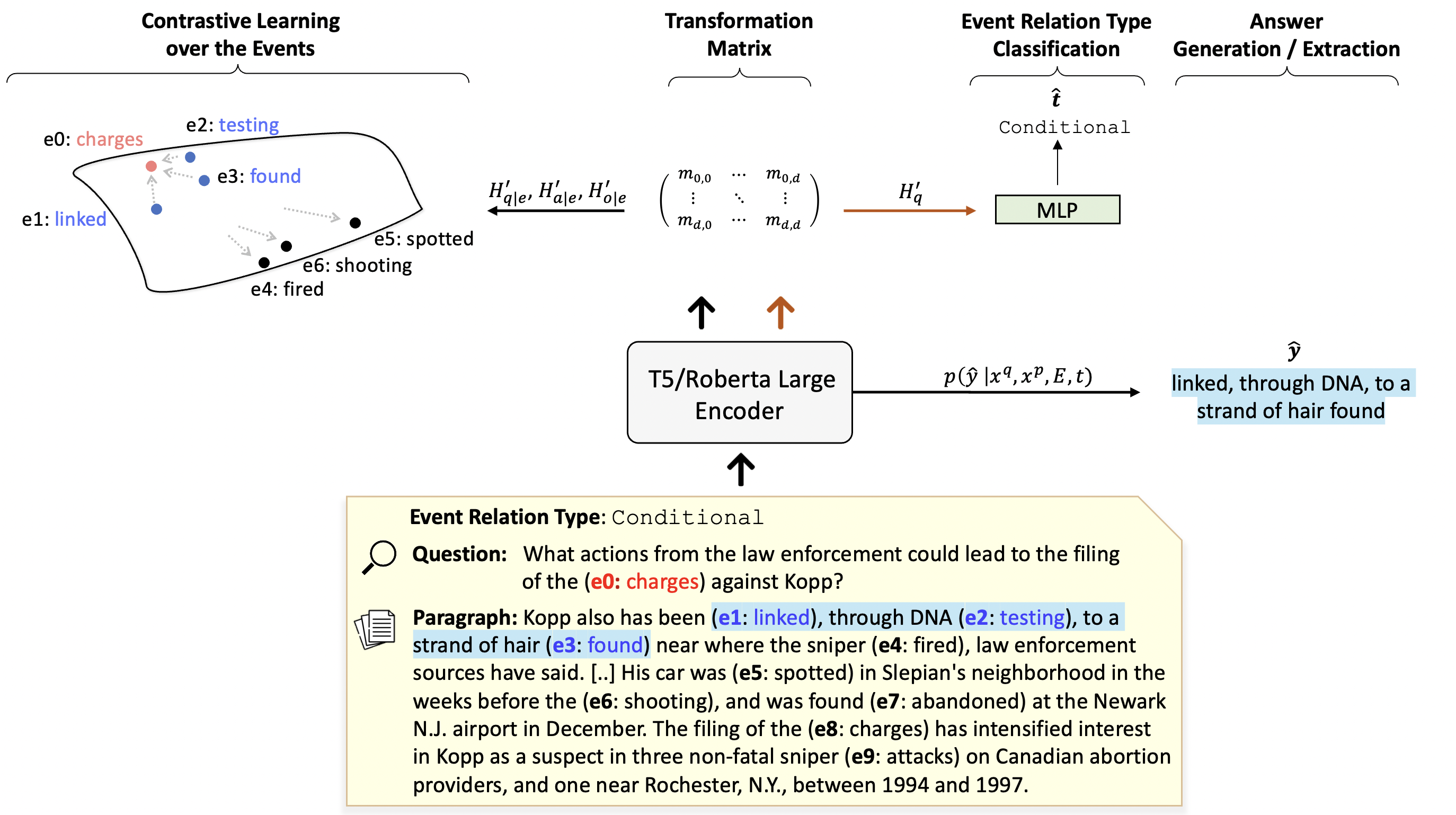}
  \caption{The overall \texttt{TranCLR} architecture. The input to encoder 
  is the concatenation of the event relation type, the question, and the paragraph. The resulting hidden vectors are used to provide answers. Simultaneously, the hidden representations are projected through a transformation matrix and used for both contrastive learning and event relation type classification. The contrastive learning mechanism realigns the event vectors to strengthen the relations between the event occurred in question and candidate answer events in paragraphs; while event relation type classification predicts the event relation type given a transformed question representation. 
  \vspace{-12pt}}
  
  \label{fig:model}
\end{figure*}

\gab{More concretely, to leverage the event semantic knowledge into QA models, we propose a novel multi-task learning framework, named \texttt{TranCLR} (Fig. \ref{fig:model}), combining a general-purpose QA model, with an event invertible transformation operator to encode event relations across questions and paragraphs. It builds on the UnifiedQA \cite{khashabi-etal-2020-unifiedqa} model for answer generation, and employs an invertible event transformation operator to project the hidden representations from the UnifiedQA encoder onto a new \textit{event embedding space}. The transformed representations are then used for (i) contrasting learning and for (ii) event relation type classification. The contrastive learning mechanism is adopted to realign the event vectors, strengthening the relations between the events mentioned in questions and those candidate answer events in paragraphs and improving the generalization to out-of-distribution event relations. On the other hand, the event relation type classification is used to further fine-tune the transformation matrix through contextualized question representations.} 
The combination of the transformation operator, along with contrastive learning and event relation type classification, leads the model to focus on the textual and relation features characterizing the event occurrences in text, and results in an overall boost in performance on event-centric QA tasks. 

Our contributions can be summarized as follows:
\gab{
\textbf{(1)} We introduce a novel multi-task learning framework for event-centric QA, \texttt{TranCLR}, in which we design an invertible event transformation operator and a contrastive learning mechanism, further combined with event relation type classification, to perform better reasoning on event semantic relations;  
\textbf{(2)} We conduct an experimental assessment on the ESTER dataset showing that \texttt{TranCLR} boosts the performance of QA models compared to strong existing PLM-based QA baselines, 
achieving over 8.4\% and 3.0\% gain in the token-level F1 and EM score respectively under the multi-answer setting;
\textbf{(3)} Visualization of event-aware token semantic vectors verifies the effectiveness of event knowledge injection. We further show the advantages of our framework tailored for event-centric learning on both zero- and few-shot learning, and adaptation ability on out-of-domain event-centric questions.
}

\section{Related work}
\gab{This work is related to two lines of research: event-centric QA, and contrastive learning.}

\paragraph{Event-centric QA} 
\gab{The growing interest into event understanding has recently led to the development of new resources for event-centric QA and event relation extraction.}
\txw{
\citet{10.1145/3340531.3412760} proposed \textit{EventQA}, an event-centric QA dataset to access semantic information stores in knowledge graphs. 
The questions are created via a random walking on the EventKG \cite{gottschalk2019eventkg}, then manually translated into natural language.
\citet{ning-etal-2020-torque} modified and converted an event temporal relation extraction dataset -- MATRES \cite{ning-etal-2018-multi} into a reading comprehension format focused on event temporal ordering questions, named TORQUE.
\gab{Instead of solely focusing on simple arguments or temporal relations, the ESTER dataset 
\cite{han2021ester} was developed to highlight how events are semantically related in terms of five most common semantic relations:  \textit{Causal, Conditional, Counterfactual, Sub-event,} and \textit{Coreference} relations.}
}
Aforementioned work built dataset baselines with popular entity-based PLMs, and thus leave significant performance gaps compared with human evaluation. \citet{asai2020logic}, \citet{dua2021learning} and \citet{shang2021open} leverage features of closely related questions to capture temporal difference to deal with certain types of event-centric questions.
\gab{Compared to the existing works, we target to various types of event-centric questions. Therefore, we introduce an invertible event transformation to (i) model the event semantic relations through an auxiliary classification task, and to (ii) realign the event latent representations via contrastive learning in the space of the transformed events.}

\paragraph{Contrastive Learning}
%
\gab{
Approaches to contrastive learning for text focus on the generation of positive and negative training pairs from 
pretrained language models.}  
\gab{
For example, \citet{clark2020electra} proposed a new pretraining framework named ELECTRA, which defines a new generative training task, i.e., Replacement Token Detection (RTD), with the aim of determining whether a token was originally replaced by the language model. Based on ELECTRA, 
\citet{meng2021coco} designed two new pretraining tasks: the Correct Language Modeling (CLM), aiming at restoring a corrupted sentence; and the a contrastive learning-based task, in which the positive pairs are made of recovered sentences and corresponding previously corrupted sentences. Similarly, \citet{qin2020erica} designed another contrastive learning framework ERICA for document-level text understanding, via specific entity
discrimination pre-training task and relation discrimination pre-training task. \citet{chen2022good} proposed a two-stage framework, integrating answer-aware span-based contrastive learning for cross-lingual machine reading comprehension. 
\citet{wu2020clear} and \citet{fang2020cert} designed a framework similar to SimCLR \cite{chen2020simple} to generate sentence representations by applying several data augmentation strategies to create contrastive pairs, such as word deleting and swapping, back-translation and synonym replacement. Yet, 
\citet{gao2021simcse} reported that  simply using dropout masks twice within a PLM can led to rather reliable positive pairs.}
\gab{Our work 
adopts standard contrastive learning framework.
The positive and negative pairs of events are composed directly from the different text sections: questions, paragraphs, and answers within the paragraph. }

\section{Methodology}
In this section, we first define the task of event-centric QA and then present our proposed TranCLR model. We build our model mainly based on the ESTER dataset \cite{han2021ester}.

\subsection{Task Formulation}
Event-centric QA can be formulated as question answering centred on the understanding of event semantic relations. 
The task can be mathematically defined as: given a text passage $\bm{x}^{p}$ 
and an answerable event-centric question $\bm{x}^q$,  
a model is asked to provide one or more answers $\bm{\hat{Y}}=\{\bm{\hat{y}}_1, \cdots, \bm{\hat{y}}_A\}$, 
where 
$A$ denotes the total number of answers to the given question. 
In the ESTER dataset, event triggers in text passages, questions and answers are annotated, $\bm{E} = \{\bm{e}_1^p,\cdots,\bm{e}_{C_p}^p,\bm{e}^q,\bm{e}_1^a,\cdots,\bm{e}_{C_a}^a\}$, where $C_p$ and $C_a$ denote the total number of events in the text passage and the answer, respectively. Each question only contains a single event $\bm{e}^q$. Since answers are parts of the paragraph in ESTER, paragraph event triggers also include answer event triggers. In addition,  
the relation type of the question event and answer events, $\bm{t}\in \mathcal{T}$, is also annotated. In the ESTER dataset, there are 
5 event semantic relations types: \emph{Causal}, \emph{Conditional}, \emph{Counterfactual}, \emph{Subevent}, and \emph{Co-reference}.

\subsection{TranCLR}
We propose a novel framework for event-centric question answering, called \texttt{TranCLR}, which is a multitask model via contrastive learning and invertible event transformation. The overall framework is shown in Figure \ref{fig:model}. Following settings in the ESTER work \cite{han2021ester}, we adopt T5-large \cite{raffel2019exploring} as an encoder-decoder backbone for 
\jr{the generative setting (i.e., answer generation), and RoBERTa-large \cite{liu2019roberta} as an encoder for the extractive setting (i.e., answer extraction).}
The T5-large model will be fine-tuned in a universal generative style \cite{khashabi-etal-2020-unifiedqa}, therefore named as UnifiedQA-T5-large. During training, the input sequence consists of question-answer event relation type label $\bm{t}$, question $\bm{x}^{q}$ and passage $\bm{x}^{p}$ with ":", "\textbackslash n", "</s>" and "<s>" special tokens. We use $\bm{x}=\{\bm{t}$:$\bm{x}^{q}\backslash n\bm{x}^{p}\}$ \jr{and $\{$<s>$\bm{t}$:$\bm{x}^{q}$</s></s>$\bm{x}^{p}\}$} to denote the whole input sequence \jr{for the generative and the extractive settings, respectively.}  
Let $N_{x}$ be the length of $\bm{x}$, and $d$ be the dimension of hidden state vectors, $H\in\mathbb{R}^{N_{x}\times d}$ is the contextual hidden states of the encoder. The target label \jr{for the generative setting} is the concatenation of all answers $\bm{\hat{Y}}=\{\bm{\hat{y}}_1, \cdots, \bm{\hat{y}}_A\}$ with each separated by a ";" special token\jr{, while labels for the extractive setting are  $\bm{\hat{Y}}=\{\bm{\hat{y}}_1, \cdots, \bm{\hat{y}}_{\bm{x}^{p}}\}$,  following the "B-I-O" or "I-O" tagging format}.

After getting the hidden states $H$ of an input sequence $\bm{x}$ via the UnifiedQA-T5-large \jr{or the RoBERTa-large} encoder, we simultaneously train the model with contrastive learning for two tasks, the main QA task, and the auxiliary task for event relation type classification. 
Therefore, our model is designed to maximize the probability $p(\bm{\hat{y}} |\bm{x}^{q},\bm{x}^{p},\bm{E},\bm{t})$ of the generated answers or \jr{the predicted labels} given a question $\bm{x}^{q}$, the supporting paragraph $\bm{x}^{p}$, all event triggers in the materials $\bm{E}$, and question-answer event relation type label $\bm{t}$. 
The event relation type $\bm{t}$ can be considered as a prefix or prompt to the input in prompt-based learning. It is worth noting that the annotated event triggers are \textbf{only} used in training, but not in inference.

The key to event-centric QA is to perform reasoning on the semantic relation of the event found in the question and those candidate events in the paired text passage. For example, for the question in Figure \ref{fig:model}, ``\emph{What actions from the law enforcemnet could lead to the filling of the charges against Kopp?}'', we would expect the QA model to generate the answer which contains the event(s) that exhibit the \emph{Conditional} relation with the event \emph{charges} mentioned in the question. This is somewhat similar to node prediction in knowledge graph embedding learning, that is, given the head event $\bm{e}^q$ in the question and the relation type $t$, we aim to locate the tail event $\bm{e}^p$ in the text passage to generate the desired answer. Inspired by 
knowledge embedding learning methods such as TransE \shortcite{bordes2013translating}, we propose to transform event embeddings using a transformation matrix and introduce an auxiliary task for event relation type classification. Using the transformation matrix has two advantages. First, the token-level hidden states are preserved in the original embedding space which are important for semantic-based QA. Second, the transformed event embeddings allow the identification of common features for more general event relation type classification in the new event embedding space. In what follows, we describe our proposed invertible event transformation operator, contrastive learning, and event relation type classification in more detail. 

\subsubsection{Invertible Event Transformation}

\gl{We propose an invertible transformation which aims to map event representations onto a new event embedding space in which the desired event semantic relations are inherently encoded.} 
Let $H_{q|e}\in\mathbb{R}^{C_{q}\times d}$, $H_{a|e}\in\mathbb{R}^{C_{a}\times d}$ and $H_{o|e}\in\mathbb{R}^{C_{o}\times d}$ be part of the hidden state vectors $H$ representing the embeddings of the question event, the answer events and other events in the text passage, in which $C_q$ refers to the number of event triggers in the question, and the sum of $C_a$ and $C_o$ refers to the total number of event triggers in the text passage. Additionally, let $M\in\mathbb{R}^{d\times d}$ be the transformation matrix, $H_{q|e}, H_{a|e}$ and $H_{o|e}$ are mapped onto a new event embedding space by: 
  $H^{'}_{q|e} = MH_{q|e}+b_{M}$,  
  $H^{'}_{a|e} = MH_{a|e}+b_{M}$, 
  and $H^{'}_{o|e} = MH_{o|e}+b_{M}$,
where $b_{M}\in\mathbb{R}^{d}$ is the bias term. \gl{The singularity of the random matrix can be guaranteed by \cite{tao2008singularity} with high probability (confirmed by our experimental results as well). Therefore, we do not need any regularisation terms to guarantee the rank of the transformation matrix in the training process.} \gl{Since the linear transformation is invertible, we have the following properties (the proof can be found in Appendix \ref{app:math}),}

\noindent  \gl{\textbf{Property 1.} For any event representation $e$ obtained from a PLM, and its transformed new embedding $e'$, we have $S(e') = S(e) + {\rm log}(|M|)$, where $S$ is the entropy of a given event.}

\gl{\textbf{Property 1} guarantees that the projected representation of a given event has a smoother distribution which makes it easier to find a separatrix in a hyperspace in the auxiliary task of event relation type classification, since $|M|$ is usually large than $1$. The distribution of outliers, i.e., low frequency words, will be smoothed by this invertible transformation as well. }

\noindent  \gl{\textbf{Property 2.} For any event representation pair $e_1$ and $e_2$ obtained from a PLM, and their transformed representations $e'_1$ and $e'_2$, we have $I(e'_1,e'_2) = I(e_1,e_2)$, where $I$ is the mutual information of the given event pair. }

\gl{\textbf{Property 2} guarantees that for any event pairs, the projection will not change the mutual information, which represents event relations encoded in the original PLM. Since the projection is a bijection and invertible, the separatrix from the learned auxiliary task will be converted to the hidden states to guide the answer generation directly. }

\subsubsection{Contrastive Learning}

After mapping event representations onto a new event embedding space by the aforementioned invertible event transformation, we can then form positive event pairs $(h_i,h_j)$ by selecting the transformed question event $h_i$ from $H^{'}_{q|e}$ and the transformed answer event $h_j$ from $H^{'}_{a|e}$. We can also form negative event pairs $(h_i,h_k)$ and $(h_k,h_j)$ by randomly sampling $h_k$ from the transformed event vectors of other events $H^{'}_{o|e}$. Let $L_{cl}$ denote the loss of contrastive learning:
\begin{equation}\small
  L_{cl} = \frac{1}{Z}\sum_{i=1}^{|H_{q|e}^{'}| + |H_{a|e}^{'}|}\sum_{j=1}^{|H_{q|e}^{'}| + |H_{a|e}^{'}|}[l_{cl:(i,j)} + l_{cl:(j,i)}]
\end{equation} 
where
  $l_{cl:(i,j)}$ denotes the loss for positive pair on event vectors $h_i$ and $h_j$, $l_{cl:(i,j)} = -\log[\exp(\mbox{cos}(h_{i}, h_{j})/\tau)/s_{i}]$, 
$\mbox{cos}(\cdot)$ denotes the cosine similarity function, $s_{i}$ denotes the sum of cosine similarity of the positive event pair $(h_{i}, h_{j})$ and that of negative event pairs $(h_i,h_{k})$, $\tau$ is the temperature hyperparameter to adjust the penalty of negative pairs, and $Z=2(|H_{q|e}^{'}| + |H_{a|e}^{'}|)$. $L_{cl}$ sums over the contrastive loss of all possible event pairs in the training set. 

\texttt{TranCLR} takes question event vector and answer event vectors as the source of positive pairs, while takes other event vectors as the source of negative events. The purpose of contrastive learning is to better employ the event information as hint for the QA task. Therefore, a good transformation matrix is essential. We introduce an auxiliary event relation type classification task in order to train a better transformation matrix. 

\subsubsection{Event Relation Type Classification}


As shown in the analysis of the $n$-gram word and token statistics conducted on the ESTER dataset \cite{han2021ester}, the questions already encode sufficient information to detect the type of event relations referred. Therefore, the idea is to apply the same transformation matrix, used on the event vectors, also on the hidden vectors encoding the question, and then use the results for  event relation type classification. 
We first predict the event relation type, $\hat{t}$ by feeding the tranformed question vector to a feed-forward layer and a softmax layer. 
We then define the cross entropy loss of event relation type classification, denoted as $L_{tc}$:
\begin{equation}
  L_{tc} =
  -\sum_{n=1}^{N}t_nlog\hat{t_n}
\end{equation}

\subsubsection{Final Objective Function}
For answer generation, the model operates on the hidden state vectors $H$ in the original embedding space encoded by UnifiedQA-T5-large (\jr{or RoBERTa-large}) to generate (or extract) the answer(s), $\bm{\hat{y}}$. 
Let $L_{qa}$ denote the loss of the main question answering task:
\begin{equation}
  L_{qa} = -\frac{1}{T}\sum_{i=1}^{T}\bm{y}_{i}\log \bm{\hat{y}_{i}}
\end{equation} 
where $T = N_a + A - 1$ is the total token length of $A$ ground truth answers separated by $A - 1$ ";" special tokens \jr{under the generative setting, while $T = \bm{x}^{p}$ is the total token length of the supporting paragraph $\bm{x}^{p}$ under the extractive setting. For the latter, we further extract all tokens marked as "BI" or "I" predictions as answers.}
The final loss is defined as:
\begin{equation}
  L = L_{qa} + \lambda_{tc}L_{tc} + \lambda_{cl}L_{cl} \label{eq:finalLoss}
\end{equation} where $\lambda_{tc}$, $\lambda_{cl}$ are hyperparameters to control the contribution of individual loss terms.

\section{Experiments}
In this section, we will first introduce the experimental setup including the dataset used and the hyperparameter setting, followed by the discussion of experimental results and ablation studies. 

\subsection{Experimental Setup}
\paragraph{Dataset} We use the event-centric QA dataset, ESTER \cite{han2021ester}, for our experiments. The dataset contains 6k human-annotated event-centric questions with an average length of 10 tokens over 1.9k paragraphs with a maximum of 340 tokens. 
All event triggers have been marked over the questions, paragraphs and answers. Besides, the dataset provides the event type label for each question from the five common event relation types: \emph{Causal}, \emph{Conditional}, \emph{Counterfactual}, \emph{Sub-event}, and \emph{Co-reference}, and collects over 10k event relation pairs. Each of the aforementioned event relation types contribute to 43.1\%, 21.3\%, 7.1\%, 15.6\% and 12.9\% of questions, respectively. Most of the questions have 1-2 in-paragraph answers, while the \emph{Sub-event} type questions have more than 3 answers on average. ESTER has been officially split into the training, development and test sets, with 4,547, 301 and 1,170 instances, respectively.\footnote{As only the training set and the development set have been released, we fine-tune our model on the training set and evaluate on the development set.} 
Table \ref{tab:dataset} in Appendix \ref{app:ester} reports the statistics of 5 event types in ESTER. 

\paragraph{Evaluation Metrics} 
We use the same metrics introduced in ESTER \cite{han2021ester}: $F_{1}^{T}$, $HIT@1$ and $EM$ defined for the multi-answer setting. $F_{1}^{T}$ calculates unigram-level token overlap between generated answers and the ground truth answers, $HIT@1$ measures whether the leftmost answer contains a correct event trigger, and Exact Match $EM$ checks if any predict answer matches exactly the corresponding ground truth answer. 

\paragraph{Baseline} The baselines we use are the seq2seq pipeline built on the UnifiedQA-T5-large \jr{and the RoBERTa-large} models introduced in ESTER \cite{han2021ester}.\footnote{As the event-centric QA task has only been recently introduced, no other approach has been proposed for ESTER.} 
Hyperparameter setting for our models can be found in Appendix \ref{app:hyper}.

\subsection{Results}

\subsubsection{Overall Comparison}

\begin{table}[ht]
\resizebox{\columnwidth}{!}{%
  \begin{tabular}{llll}
    \toprule
    Model & $F_{1}^{T}$ & $HIT@1$ & $EM$\\
    \midrule
    \multicolumn{4}{c}{\emph{Generative setting}}\\
    \midrule
    UnifiedQA-large \cite{han2021ester} & 66.8 & \textbf{87.2} & 24.4 \\
    UnifiedQA-large (our run) & 65.8 & 86.7 & 24.6 \\
    UnifiedQA-large TranCLR & \textbf{74.2} & 86.4 & \textbf{25.6} \\
    \midrule
    UnifiedQA-large TranCLR (-prefix) & 69.6 & 81.4 & 21.6 \\
    UnifiedQA-large TranCLR (-TC) & \textbf{74.6} & \textbf{87.4} & 24.6 \\
    UnifiedQA-large TranCLR (-CL) & 72.8 & 84.7 & \textbf{25.6} \\
    UnifiedQA-large TranCLR (-TransM) & 66.8 & 77.7 & 20.3 \\
    \midrule
   \multicolumn{4}{c}{\emph{Extractive setting}}\\
    \midrule
    RoBERTa-large \cite{han2021ester} & 68.8 & 66.7 & 16.7 \\
    RoBERTa-large (our run) & 67.0 & 69.4 & 17.9 \\
    RoBERTa-large (IO) & 73.7 & 77.4 & 15.3 \\
    RoBERTa-large (IO) TranCLR & \textbf{74.7} & \textbf{80.4} & \textbf{18.3} \\
    \bottomrule
  \end{tabular}}
    \caption{Main results of experiments on ESTER dataset. \cite{han2021ester} takes "B-I-O" labels for extractive QA, while we found "I-O" labels work better. UnifiedQA-large TranCLR (-*) refer to ablation studies for generative QA, where -prefix, -TC, -CL and -TransM refer to the removal of the event type label prefix, question-answer event type classification, contrastive learning, and the transformation matrix, respectively.}
  \label{tab:twosettings}
\end{table}

\begin{table*}[t]
\resizebox{\textwidth}{!}{%
  \begin{tabular}{l|ccc|ccc|ccc|ccc}
    \toprule
    \quad & \multicolumn{3}{c|}{UnifiedQA-large (our run)} & \multicolumn{3}{c|}{UL TranCLR} & \multicolumn{3}{c|}{UL TranCLR (-TC\&CL)} & \multicolumn{3}{c}{UL TranCLR (-prefix)}\\
    \toprule
    Type & $F_{1}^{T}$ & $HIT@1$ & $EM$ & $F_{1}^{T}$ & $HIT@1$ & $EM$ & $F_{1}^{T}$ & $HIT@1$ & $EM$ & $F_{1}^{T}$ & $HIT@1$ & $EM$\\
    \midrule
    Causal (39.2\%) & 72.8 & \textbf{90.7} & \textbf{31.4} & \textbf{80.5} & 89.8 & 30.5 & 71.4 & 88.1 & \textbf{31.4} & 75.0 & 82.2 & 25.4\\
    Conditional (19.3\%) & 58.7 & 84.5 & 19.0 & 66.6 & 86.2 & \textbf{24.1} & 63.5 & \textbf{89.7} & 22.4 & \textbf{69.2} & 86.2 & 19.0\\
    Counterfactual (9.3\%) & 65.7 & \textbf{78.6} & 35.7 & \textbf{75.9} & \textbf{78.6} & 32.1 & 65.1 & 75.0 & \textbf{39.3} & 65.2 & 67.9 & 28.6\\
    Sub-event (19.6\%) & 59.0 & 89.8 & \textbf{13.6} & \textbf{70.6} & 89.8 & \textbf{13.6} & 66.2 & \textbf{93.2} & \textbf{13.6} & 65.0 & 89.8 & 11.9\\
    Co-reference (12.6\%) & 65.2 & \textbf{79.0} & 21.1 & \textbf{70.6} & 76.3 & \textbf{26.3} & 65.8 & 76.3 & 23.7 & 63.6 & 68.4 & 23.7\\
    \bottomrule
    All (100\%) & 65.8 & \textbf{86.7} & 24.6 & \textbf{74.2} & 86.4 & 25.6 & 67.6 & \textbf{86.7} & \textbf{25.9} & 69.6 & 81.4 & 21.6\\
    \bottomrule
  \end{tabular}}
    \caption{Results from various models on 5 different event relation types on the development set. UL refer to the abbreviation of UnifiedQA-large. UL \texttt{TranCLR} outperforms UnifiedQA-large baseline significantly in $F_1^T$ across all event relation types. The ablated versions of UL \texttt{TranCLR} show mixed results in $HIT@1$ and $EM$. }
  \label{tab:typeresults}
\end{table*}

We show the overall evaluation results in Table \ref{tab:twosettings}. Our model achieves impressive results compared with the previous baseline, gaining about \jr{8\% improvement in $F_{1}^{T}$ under the generative setting, 3.0\% improvement in $EM$ and $HIT@1$ scores under the extractive setting.\footnote{Evaluation code: \url{https://github.com/PlusLabNLP/ESTER/tree/master/code}}}
We have the following observations: (1) event-based contrastive learning brings a significant gain, enabling a better reasoning of semantic relations between event triggers in questions and candidate answers in text, since the question event and the answer event, although bearing very different semantic meanings, are pushed closer in their projected new event embedding space. This is evident from the drastically improved  $F_{1}^{T}$ score in the main QA task; (2) both event relation type classification and contrastive learning are indispensable since the combination of them achieves more balanced results across all metrics in answer evaluation, showing that the auxiliary event relation type classification task leads to a better learned transformation matrix; (3) prompt-based learning using the event relation type as prefix of input\footnote{During inference, event relation type is detected automatically from a given question.} is effective as the additional information can better guide the model to answer questions which are much more difficult than traditional factoid questions; and finally (4) the use of the transformation matrix makes it possible to simultaneously learn representations in both the original embedding space and the new event-centric space, leading to better QA results.

\subsubsection{Zero-shot and Few-shot Learning}

\begin{figure}[ht]
  \centering
  \includegraphics[width=\linewidth]{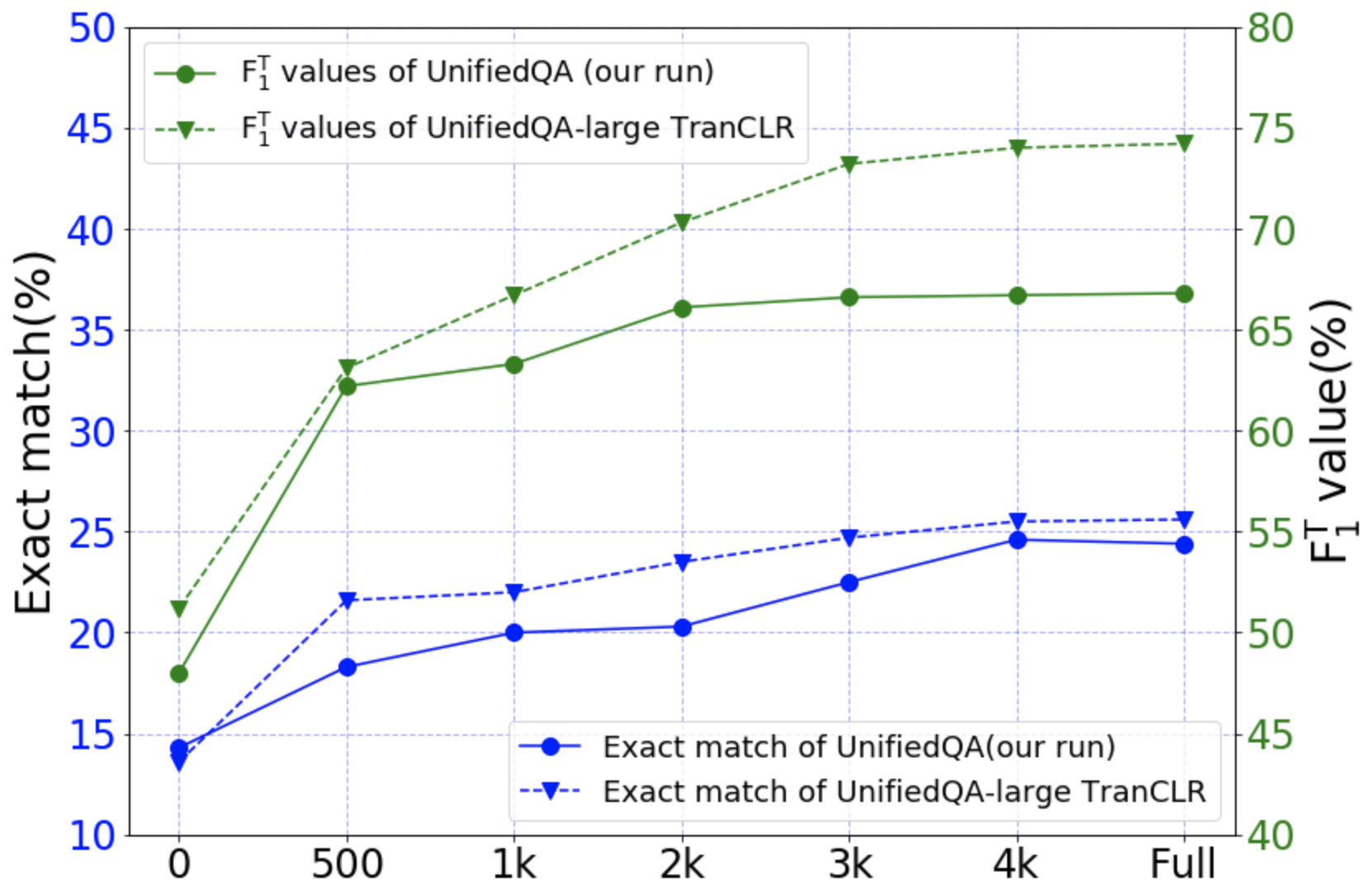}
  \caption{Zero- and Few-shot learning performance of UnifiedQA-large TranCLR and UnifiedQA-large. 
  }
  \label{fig:few-shot}
\end{figure}

In this section, we assess the ability of \texttt{TranCLR} for zero-shot and few-shot learning, i.e., without the training data or with very few training instances. Figure \ref{fig:few-shot} shows the $F_1^T$ and $EM$ values of \texttt{TranCLR} and the baseline UnifiedQA-large with the varying size of the training set. It can be observed that in zero-shot learning, both models give similar $EM$ results and \texttt{TranCLR} slightly outperforms UnifiedQA-large in $F_1^T$. With only 500 training instances, \texttt{TranCLR} is able to generate more accurate answers, beating UnifiedQA-large by 3\% in $EM$, demonstrating the benefit of making effective use of event information for better reasoning of event semantic relations. 
The performance gap however gradually diminishes with the increasing size of the training set. Nevertheless, \texttt{TranCLR} is able to generate answers containing more overlapped information with the ground truth with more training instances, evidenced by the increased performance gains compared to UnifiedQA-large, reaching nearly 8\% in $F_1^{T}$ when using the full training set. This shows that our proposed contrastive learning combined with the auxiliary task of event semantic type classification can better capture event semantic knowledge which guides the decoder to generate answers closer to the ground truth.

\begin{table*}[t]
\resizebox{\textwidth}{!}{%
  \begin{tabular}{l}
    \toprule
    \textbf{Paragraph}: Kopp also has been \textbf{linked, through DNA testing, to a strand of hair found} near where the sniper fired, law enforcement sources have\\ said.\textbackslash n Nicknamed the ``Atomic Dog'' in anti-abortion circles, Kopp had been arrested in several states since 1990 for protesting abortion. His car was\\ spotted in Slepian's neighborhood in the weeks before the shooting, and was found abandoned at the Newark, N.J., airport in December.\textbackslash n The filing of\\ the charges has intensified interest in Kopp as a suspect in three non-fatal sniper attacks on Canadian abortion providers, and one near Rochester, N.Y.\\, between 1994 and 1997.\textbackslash n Kopp is now the second anti-abortion activist being sought by the FBI as a suspect in a fatal attack.\textbackslash n \\
    \midrule
    \textbf{Question}: What actions from the law enforcement could lead to the filing of the charges against Kopp?\\
    \textbf{Ground Truth Answer}: linked, through DNA testing, to a strand of hair found\\
    \textbf{Question-answer event relation type}: \emph{Conditional}\\
    \midrule
    \textbf{UnifiedQA-T5-large (our run)}: 1: arrested in several states since 1990 for protesting abortion; 2: his car was spotted in slepian's\\ neighborhood; 3: was found abandoned at the newark, n.j., airport in december\\
    \textbf{UnifiedQA-T5-large TranCLR}: 1: arrested in several states since 1990; 2: link, through dna testing; 3: strand of hair found near where the sniper fired\\
    \bottomrule
  \end{tabular}}
    \caption{Example answers generated by different models. \texttt{TranCLR} injected with event knowledge accurately grab the news narrative, and generate answers that cover 100\% content of ground truth. In contrast, UnifiedQA-T5 without event-related learning is confused with question-related context information in the paragraph.}
  \label{tab:generatedcase}
\end{table*}

\subsubsection{Results per Event Relation Type} In Table \ref{tab:typeresults}, we provide detailed comparison of results under various event relation types. We can observe that the results on the `\emph{Causal}' type, being the largest category, are much better compared to other event relation types. Our proposed \texttt{TranCLR} achieves the best $F_1^T$ scores across all event relation types compared to the baseline UnifiedQA-large, with the increment in the range of 5.4-11.6\%. The largest performance improvement of 11.6\% is observed on the most difficult `\emph{Sub-event}' type in which questions have more than 3 answers on average. 
By analyzing the results, we found that it is sometimes quite difficult to distinguish between `\emph{Conditional}' and `\emph{Counterfactual}' types. As such, adding the event relation type as prefix may confuse the model. In terms of the $HIT@1$ results, 
\texttt{TranCLR} with prefix only (i.e., \textsf{-TC\&CL}) improves upon the baseline by over 5\% and nearly 4\% for the `\emph{Conditional}' and the `\emph{Sub-event}' types respectively. We also observe that using the event relation type as prefix in prompt-based learning is very effective in boosting the $EM$ scores, especially for the `\emph{Counterfactual}' type in which nearly 4\% improvement is obtained compared to UnifiedQA-large. 
For the `\emph{Sub-event}' type where multiple answers are expected, there is no improvement in $EM$ in our models compared to the baseline.

\subsubsection{Visualisation of Event Embeddings} 

\begin{figure}[t]
  \centering
  \includegraphics[width=\linewidth]{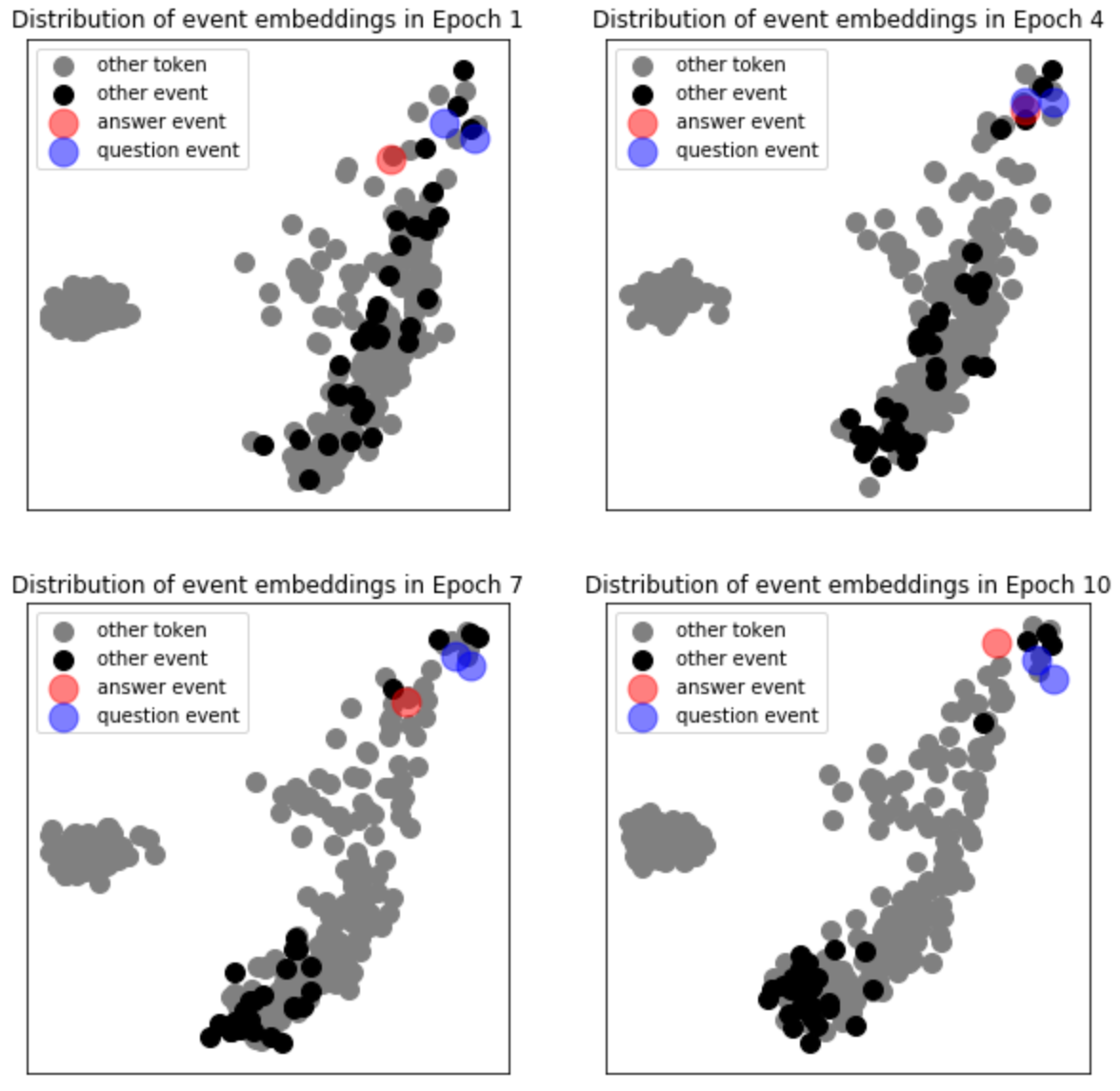}
  \caption{Distributions of events in the semantic space. Question events, answer events, other events and the remaining non-event tokens are shown in blue, red, black and gray, respectively. 
  }
  \label{fig:visualization}
\end{figure}

In Figure \ref{fig:visualization}\footnote{Better viewing in color.}, 
we visualize the learned event embeddings in the semantic space during the training. 
It can be observed that with the increasing number of training epochs, event triggers are grouped into few clusters from an evenly distributed initial state. Most irrelevant event nodes are pushed aside as they are used as negative samples in contrastive learning, while question events and answer events are pulled together. The visualization reveals reasonable process of event knowledge distillation.

\subsection{Qualitative Analysis}
We further perform qualitative analysis in Table \ref{tab:generatedcase} over the example illustrated in Figure \ref{fig:sample} and \ref{fig:model}\footnote{More example outputs are presented in Appendix \ref{app:morecase}.}. All models generate more than one answers. 
More concretely, \texttt{TranCLR} 
manages to generate text covering 100\% of the ground truth answer, while UnifiedQA-large is unable to generate coherent answers as the model failed to detect the semantic relations 
between the question event trigger "\emph{charges}" and the answer event triggers "\emph{link}", "\emph{testing}" and "\emph{found}". 

\subsection{Generalization Evaluation}

To further evaluate the generalization capability of our event knowledge distillation paradigm, we apply the \texttt{TranCLR} model trained on ESTER for zero-shot inference on unseen QA data focusing on event temporal relations, using the TORQUE dataset \cite{ning2020torque}. TORQUE focuses on questions about event temporal relations such as \emph{“what happened before/after [some event]?”}. For each question, the dataset provides a two-sentence supporting passage and passage event annotations. The answers are simply event mentions in the form of words/phrases, rather than longer text spans as in the ESTER dataset. We perform zero-shot inference on TORQUE without fine-tuning the models on its training set. It can be observed from Table \ref{tab:TORQUE} that compared with RoBERTa-large without trained on any event QA data (first result row), fine-tuning RoBERTa-large on ESTER (second result row) improves $F_{1}$ by 10\%. Nevertheless, our proposed \texttt{TranCLR} exhibits a significantly better event understanding capability, achieving 8.7\% and 11.5\% further gains in $F_{1}$ and EM scores, respectively. It is worth mentioning that the ESTER dataset does not contain any questions about event temporal relations. The results show the strong generalization capabilities of \texttt{TranCLR} and further verify the effectiveness of our proposed framework for event semantic reasoning.

\begin{table}[t]
\resizebox{\columnwidth}{!}{%
  \begin{tabular}{lll}
    \toprule
    Model & $F_{1}$ & $EM$\\
    \midrule
    RoBERTa-large & 10.0 & 0.0\\
    RoBERTa-large \cite{han2021ester} & 20.0 & 4.1\\
    RoBERTa-large TranCLR (ESTER) & \textbf{28.7} & \textbf{15.6}\\
    \bottomrule
  \end{tabular}}
    \caption{Zero-shot inference results on the TORQUE development set. 
    RoBERTa-large (the first result row) is not trained on any event QA data, while the other models are trained on ESTER only.}
  \label{tab:TORQUE}
\end{table}

\section{Conclusions}
In this paper, we have proposed a novel framework, called \texttt{TranCLR}, to tackle the event-centric QA task on the ESTER dataset \cite{han2021ester}. The core idea of \texttt{TranCLR} is to effectively explore the event knowledge in both questions and context through event-centric contrastive learning and the auxiliary task of event type classification. 
Our experimental results show superior performance of \texttt{TranCLR} on event-centric QA compared to the strong baseline, gaining 8.4\% and 3\% absolute improvements in $F_1^{T}$ and EM scores respectively. \jr{Further zero-short inference and qualitative analysis verify the promising event semantic understanding and reasoning capability of our model.}

\section*{Limitations} 

Although we have verified the promising event semantic understanding and reasoning capability of \texttt{TranCLR} trained on ESTER for both in-domain event semantic relations and the out-of-domain event temporal relation, it is worth further exploring whether the model indeed captures event semantic relations and does not just generate answers by the matching of spurious patterns. Adversarial attacks could be explored in the future to assess the possible backdoor of the model in order to evaluate its robustness \cite{hyper21, disen21, bartolo2021improving}.

Our current work is built on the ESTER dataset where each question is paired with a single paragraph. In reality, event-centric QA may require the gathering of evidence scattered over multiple paragraphs and reasoning over more sophisticated event chains or graphs. Such complex event semantic relations is beyond what our proposed event-centric contrastive learning could capture. To develop new methodologies for dealing with more challenging event-centric QA, efforts need to be devoted to develop a dataset under a more realistic setting.



\section*{Acknowledgement}

This work was funded in part by the the UK Engineering and Physical Sciences Research Council (grant no. EP/T017112/1, EP/V048597/1, EP/X019063/1). YH is supported by a Turing AI Fellowship funded by the UK Research and Innovation (grant no. EP/V020579/1). The work has been conducted on the UKRI/EPSRC HPC platform, Avon, hosted in the University of Warwick’s Scientific Computing Group.

\bibliography{emnlp2022}
\bibliographystyle{acl_natbib}

\clearpage
\appendix

\setcounter{table}{0}
\renewcommand{\thetable}{A\arabic{table}}

\section*{Appendix}
\section{Proof of Properties of the Invertible Transformation}
\label{app:math}
\noindent \gl{\textbf{Definition:} For any event representation $e$ obtained from a Pre-trained Language Model (PLM), let $e'$ be the transformed representation, $e' = M \cdot e + b$, where $M$ is the transformation matrix and $b$ is a bias, the projection of linear transformation is invertible. }

\noindent  \gl{\textbf{Property 1} For any event representation $e$ obtained from a PLM, and its transformed representation $e'$, we have $S(e') = S(e) + {\rm log}(|M|)$, where $S$ is the entropy of the given event.}

\gl{\textbf{Proof.} Assume that $M$ is an identity matrix, we then have $e' = e + b$. Hence,}

{\small
\begin{equation}
\begin{aligned}
    S(e') &= -\int p_{e'}(e + b) {\rm log} p_{e'}((e + b) d(e+b) \\
    &= -\int p_e(e) {\rm log} p_e(e) d(e) \\
    &= S(e),
\end{aligned}
\end{equation}}

\gl{\noindent where the $p_e$ and $p_{e'}$ represent the probability space before and after transformation. Therefore, the bias term will not change the entropy after projection. Then, we only need to consider a general transformation matrix $M$,}

{\small
\begin{equation}
\begin{aligned}
    S(e') &= -\mathbb{E}[{\rm log}p_{e'}(M \cdot e)] \\
    &= -\mathbb{E}[{\rm log}(M^{-1}p_e(M^{-1} \cdot e'))] \\
    &= -\mathbb{E}[{\rm log}(M^{-1}p_e(e)] \\
    &= S(e) + {\rm log}(|M|)\\ 
\end{aligned}
\end{equation}}
\qedsymbol

\noindent  \gl{\textbf{Property 2} For any event representation pair $e_1$ and $e_2$ obtained from a PLM, and their transformed representations $e'_1$ and $e'_2$, we have $I(e'_1,e'_2) = I(e_1,e_2)$, where the $I$ is the mutual information of the given event pair. }

\gl{\textbf{Proof.} }

{\small
\begin{equation}
\begin{aligned}
I(e'_1,e'_2) &= S(e'_1) - S(e'_1|e'_2) \\
&= S(e'_1) - S(e'_1|e'_2) +  {\rm log}(|M|)- {\rm log}(|M|)
\end{aligned}
\end{equation}}

\gl{According to the proof of \textbf{Property 1}, we have $S(e) = S(e') - {\rm log}(|M|)$. Hence, }

{\small
\begin{equation}
\begin{aligned}
I(e'_1,e'_2) &= S(e'_1) - {\rm log}(|M|) - (S(e'_1|e'_2) - {\rm log}(|M|)) \\
& = S(e_1) - S(e_1|e_2) \\
&= I(e_1, e_2)
\end{aligned}
\end{equation}}
\qedsymbol

\section{Experimental Setting}

\paragraph{Hyperparameters}
\label{app:hyper}
Our hyperparameter setting follows what has been reported in \citet{han2021ester}. 

For generative setting, the hidden size of Unified-T5-large is 1,024 and the corresponding vocabulary size is 32,128. The random seed is 5. The batch size is set to 2 and the accumulation steps 3 on 2 quadro\_rtx\_6000 GPUs. The optimizer of all models is BertAdam\footnote{\url{https://github.com/google-research/bert/blob/master/optimization.py}} with $\beta1=0.9, \beta2=0.999$, and $\epsilon$=1e-6. Except for parameters of weights in layer normalization and bias in all layers, all other trainable parameters are decayed with a rate of 0.95 during training. The learning rate is increased linearly from 0 to 5e-5 in the first 10\% total training steps and then linearly decreased to 0. 

Similarly, for extractive setting, the hidden size of Roberta-large is 1,024 and the vocabulary size is 50,265. The random seed is 23. The batch size is 8 with accumulation steps 2 on same 2 quadro\_rtx\_6000 GPUs. The optimizer and decaying strategy remain same as generative models. The learning rate is changed to 1e-5. In addition, following \citet{han2021ester}, we adopt label weight 4 for "B" and "I" label to reduce label unbalance.

For other hyperparamters, we set empirically $\tau$ to 1.0 in contrastive learning, and $\lambda_{tc}$ = 0.1, $\lambda_{cl}$ = 0.1 in Eq. (\ref{eq:finalLoss}). It takes around 3 hours to fine-tune our models for 10 epochs. Parameter amounts are 356M and 738M for extractive and generative settings respectively.

\paragraph{Statistics of the ESTER Dataset}
\label{app:ester}

Table \ref{tab:dataset} shows the data statistics per event relation type. The \emph{Causal} event relation type is the largest category, while \emph{Counterfactual} being the smallest one.

\begin{table}[h!]\small
  \centering
  \begin{tabular}{llll}
    \toprule
    Type & Train & Dev & Test\\
    \midrule
    Causal & 2047 & 118 & 431 \\
    Conditional & 928 & 58 & 289 \\
    Counterfactural & 294 & 28 & 106 \\
    Sub-event & 678 & 59 & 204 \\
    Co-reference & 600 & 38 & 140 \\
    \midrule
    All & 4547 & 301 & 1170 \\
    \bottomrule
  \end{tabular}
    \caption{The statistics of 5 event types in the ESTER dataset. We only use the training and development set. The test set is not published.}
  \label{tab:dataset}
\end{table}

\section{More Generated Answers}
\label{app:morecase}

\begin{table*}[t]
\resizebox{\textwidth}{!}{
  \begin{tabular}{l}
    \toprule
    \textbf{Paragraph}: Another leftist South American nation, Bolivia, has also expressed a desire to join OPEC despite its modest oil production of\\ 40,000 barrels per day.\textbackslash n The short-term outlook in terms of OPEC's influence on oil prices is likely to depend on the discipline of the cartel\\ and the actual reduction of supplies to the market.\textbackslash n The cartel decided in October to reduce its output by 1.2 million bpd from the\\ beginning of November, but analysts believe the real reduction has been only 500,000-800,000 bpd because of cheating by some members.\\ \textbackslash n The cut of 500,000 bpd in February would reduce the output from OPEC members, excluding Iraq and Angola, to 25.8 million bpd in\\ principle.\textbackslash n Some analysts expressed concern that any reduction in supplies could send prices higher in the months ahead, the peak time\\ for \textbf{oil demand because of the northern hemisphere winter}.\textbackslash n\\
    \textbf{Question}: Why could a reduction in supplies send prices higher in the months ahead?\\
    \textbf{Answer 1}: oil demand because of the northern hemisphere winter\\
    \textbf{Question-answer event relation type}: Conditional\\
    \textbf{UnifiedQA-T5-large(our run)}: 1: the peak time for oil demand because of the northern hemisphere winter\\
    \textbf{UnifiedQA-T5-large TranCLR}: 1: the peak time for oil demand because of the northern hemisphere winter\\
    \bottomrule
    \bottomrule
    \textbf{Paragraph}: Lowe's decision to bring former England rugby world cup-winning coach Clive Woodward into the backroom staff was one\\ of the reasons for his unhappiness.\textbackslash n "As a bloke I got on with him but I have to say \textbf{the decision to bring him in was bizarre}. \textbf{The}\\ \textbf{relationship between me and Clive was never going to work} because there were too many people undermining the structure Rupert\\ Lowe wanted at the club," he added.\textbackslash n In an open letter by Mandaric published in the Sunday Mirror the Pompey chairman said: "If I'm\\ honest I \textbf{never wanted Harry to leave in the first place}.\textbackslash n "Of all the candidates Harry is the one which that stands out. The supporters\\ have to trust me."\textbackslash n\\
    \textbf{Question}: Why was Mandaric unhappy with Lowe bringing Clive onto the staff?\\
    \textbf{Answer 1}: never wanted Harry to leave in the first place\\
    \textbf{Answer 2}: The relationship between me and Clive was never going to work\\
    \textbf{Answer 3}: the decision to bring him in was bizarre\\
    \textbf{Question-answer event relation type}: Causal\\
    \textbf{UnifiedQA-T5-large(our run)}: 1: too many people undermining the structure rupert lowe wanted at the club\\
    \textbf{UnifiedQA-T5-large TranCLR}: 1: the decision to bring him in was bizarre\\
    \bottomrule
    \bottomrule
    \textbf{Paragraph}: The visiting U.S. Assistant Secretary of State Richard Boucher on Tuesday said the United States did not have any\\ involvement in the attack on a religious school in Pakistan's tribal region.\textbackslash n "The \textbf{Pakistani government has said they carried out}\\ initiative to deal with serious threats from fighters who were in that location," Boucher told reporters at U.S. embassy.\textbackslash n "The \textbf{Pakistani}\\ \textbf{government says it has carried out this action}. And it was necessary because militants, terrorists created a training center," he said.\\\textbackslash n He supported Pakistan's policy to engage tribal elders to establish peace in the tribal region. \textbackslash n\\
    \textbf{Question}: What might have made Richard Boucher say United States did not have any involvement in the attack?\\
    \textbf{Answer 1}: Pakistani government has said they carried out\\
    \textbf{Answer 2}: Pakistani government says it has carried out this action\\
    \textbf{Question-answer event relation type}: Conditional\\
    \textbf{UnifiedQA-T5-large(our run)}: 1: the pakistani government has said they carried out initiative to deal with serious threats\\
    \textbf{UnifiedQA-T5-large TranCLR}: 1: the pakistani government has said they carried out initiative\\
    \bottomrule
    \bottomrule
    \textbf{Paragraph}: It was Chamara Silva who primarily kept the scoreboard ticking over with a 68-run stand for the fifth wicket with Mahela\\ Jayawardene, and an unbeaten 57-run partnership with Prasanna Jaywardene.\textbackslash n At the close, Silva was unbeaten on 79, his second half\\ century of the match after failing to score in the first Test, while Jayawardene was not out 22.\textbackslash n Lasith "Slinga" Malinga and Muttiah\\ Muralitharan had earlier bowled Sri Lanka to a 138-run first innings lead.\textbackslash n The New Zealand batsmen had no answer to the hostile pace\\ and slinging action of Malinga at one end, and could not read Muralitharan's spin at the other, as they crumbled to be all out before lunch\\ on the second day for 130.\textbackslash n Of the New Zealand batsmen only Brendon McCullum put up any solid resistance. He was dropped on the\\ first ball of the day without scoring and went on to post 43 before he was bowled by Muralitharan to end the innings.\\
    \textbf{Question}: What led to the crumbling of New Zealand on the second day?\\
    \textbf{Answer 1}: hostile pace and slinging action of Malinga\\
    \textbf{Answer 2}: could not read Muralitharan's spin\\
    \textbf{Question-answer event relation type}: Causal\\
    \textbf{UnifiedQA-T5-large(our run)}: 1: hostile pace and slinging action of malinga\\
    \textbf{UnifiedQA-T5-large TranCLR}: 1: hostile pace and slinging action of malinga at one end; 2: could not read muralitharan's spin\\
    \bottomrule
    \bottomrule
    \textbf{Paragraph}: Heavy fighting resumed in central Somalia Wednesday after retreating Islamist fighters opened fire on a force of government\\ and Ethiopian troops, officials and residents said, as the conflict in the Horn of Africa nation entered its second week.\textbackslash n Hours after the\\ UN Security Council failed to agree on the withdrawal of foreign troops, \textbf{Islamists fighters in trenches near the town of Jowhar}\\ \textbf{opened fire} to stop Somali-Ethiopian troops from advancing further southwards.\textbackslash n "Very heavy fighting has erupted outside Jowhar.\\ The Islamic forces say they will keep fighting," said Mohamed Abdi Ali, a resident of the town about 90 kilometres (55 miles) north of the\\ Islamist-controled capital of Mogadishu.\textbackslash n "Ethiopians have not started using planes yet, but we do not rule that out," he added.\textbackslash n\\
    \textbf{Question}: What could be expected to happen after the Somali-Ethiopian troops tried to advance southwards?\\
    \textbf{Answer 1}: Islamists fighters in trenches near the town of Jowhar opened fire\\
    \textbf{Question-answer event relation type}: Conditional\\
    \textbf{UnifiedQA-T5-large(our run)}: 1: heavy fighting resumed in central somalia\\
    \textbf{UnifiedQA-T5-large TranCLR}: 1: Islamists fighters in trenches\\
    \bottomrule
  \end{tabular}}
    \caption{Additional generated samples from the selected models. In the first case, both models generate overlong answers. In the second case, \texttt{TranCLR} manages to generate one completely correct answer while UnifiedQA-T5-large produced a wrong answer. For the next two cases, \texttt{TranCLR} controls answer range better in the third one and is able to cover both answers in the fourth one, compared with the UnifiedQA-T5-large baseline. In the last case, only \texttt{TranCLR} generates related answer.}
  \label{tab:morecases}
\end{table*}

\end{document}